\documentclass[conference]{IEEEtran}
\IEEEoverridecommandlockouts
\usepackage{cite}
\usepackage{amsmath,amssymb,amsfonts}
\usepackage{algorithmic}
\usepackage{graphicx}
\usepackage{textcomp}
\usepackage{xcolor}

\usepackage[hidelinks]{hyperref}

\usepackage[linesnumbered,ruled]{algorithm2e}
\def\BibTeX{{\rm B\kern-.05em{\sc i\kern-.025em b}\kern-.08em
    T\kern-.1667em\lower.7ex\hbox{E}\kern-.125emX}}
\begin{document}

\title{
\huge{SuperDriverAI: Towards Design and Implementation for\\End-to-End Learning-based Autonomous Driving
}
}


\author{
Shunsuke Aoki$^{1,2}$,
Issei Yamamoto$^{2}$,
Daiki Shiotsuka$^{2}$,
Yuichi Inoue$^{2}$,
Kento Tokuhiro$^{2}$,
and Keita Miwa$^{2}$
\thanks{$^{1}$ Shunsuke Aoki is with the National Institute of Informatics, Japan.}
\thanks{$^{2}$ Shunsuke Aoki, Issei Yamamoto, Daiki Shiotsuka, Yuichi Inoue, Kento Tokuhiro, and Keita Miwa are with TURING, Inc., Japan.}
}
\maketitle

\begin{abstract}
Fully autonomous driving has been widely studied and is becoming increasingly feasible.
However, such autonomous driving has yet to be achieved on public roads, because of various uncertainties due to surrounding human drivers and pedestrians.
In this paper, we present an end-to-end learning-based autonomous driving system named \textit{SuperDriver AI}, where Deep Neural Networks (DNNs) learn the driving actions and policies from the experienced human drivers and determine the driving maneuvers to take while guaranteeing road safety.
In addition, to improve robustness and interpretability, we present a \textit{slit model} and a visual attention module.
We build a data-collection system and emulator with real-world hardware, and we also test the SuperDriver AI system with real-world driving scenarios.
Finally, we have collected $150$ runs for one driving scenario in Tokyo, Japan, and have shown the demonstration of SuperDriver AI with the real-world vehicle.

\end{abstract}


\section{Introduction}
Since the DARPA Urban Challenge for autonomous driving in 2007 \cite{urmson2008autonomous}, many researchers and engineers have worked for autonomous driving systems.
For example, Carnegie Mellon University has outfitted a Cadillac SRX to drive itself and has also developed a tool-chains to aid the testing of autonomous driving systems \cite{bhat2018tools}.
In fact, there are multiple commercial products for autonomous driving, made by Tesla and comma.ai, and traditional automakers have offered driver-assistance systems.
Although autonomous driving technologies have been deployed into the real world, fully autonomous driving has yet to be achieved on public roads, due to the two challenges: (i) Cooperation with the surrounding humans and (ii) Unexpected road scenarios.
First, autonomous driving systems might be difficult to determine their behaviors in front of human drivers and/or pedestrians \cite{aoki2021commmagazine}. Typically, there are traffic rules to guarantee road safety on public roads, but human drivers and pedestrians may violate the traditional traffic rules for operational purposes.
Secondly, due to construction, traffic control, blocking obstacles, and/or potholes, roads might have \textit{dynamic intersection} \cite{aoki2018dynamic}, in which unexpected road scenarios might happen and each driver might require to cooperate with by using combination of traffic rules, culture, courtesy, social norms, and hand gestures.
Hence, autonomous driving systems navigate themselves under such dynamic situations while understanding the explicit and implicit driving behaviors of the surrounding human drivers.


\begin{figure}[!b]
  \begin{center}
    \begin{tabular}{c}
      \begin{minipage}{0.5\hsize}
        \begin{center}
          \includegraphics[clip, width=2.5cm]{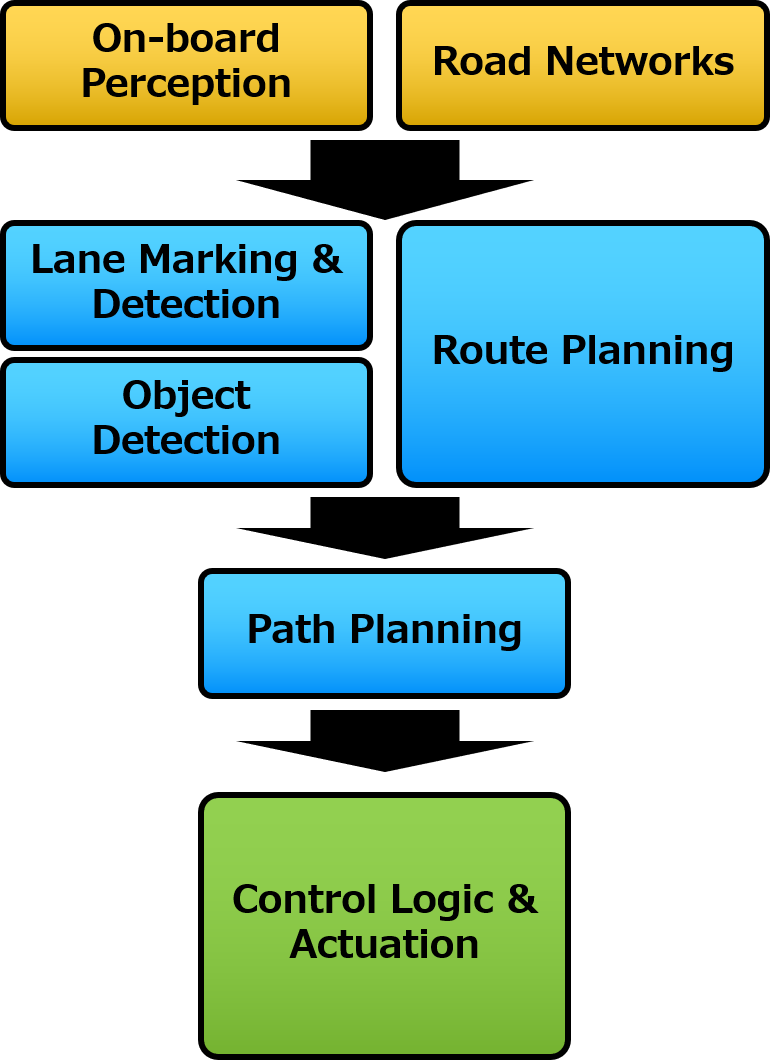}
	\hspace{2.6cm} (a) Traditional Approach.
        \label{fig:Synchro_2size}
        \end{center}
      \end{minipage}

      \begin{minipage}{0.5\hsize}
        \begin{center}
          \includegraphics[clip, width=2.5cm]{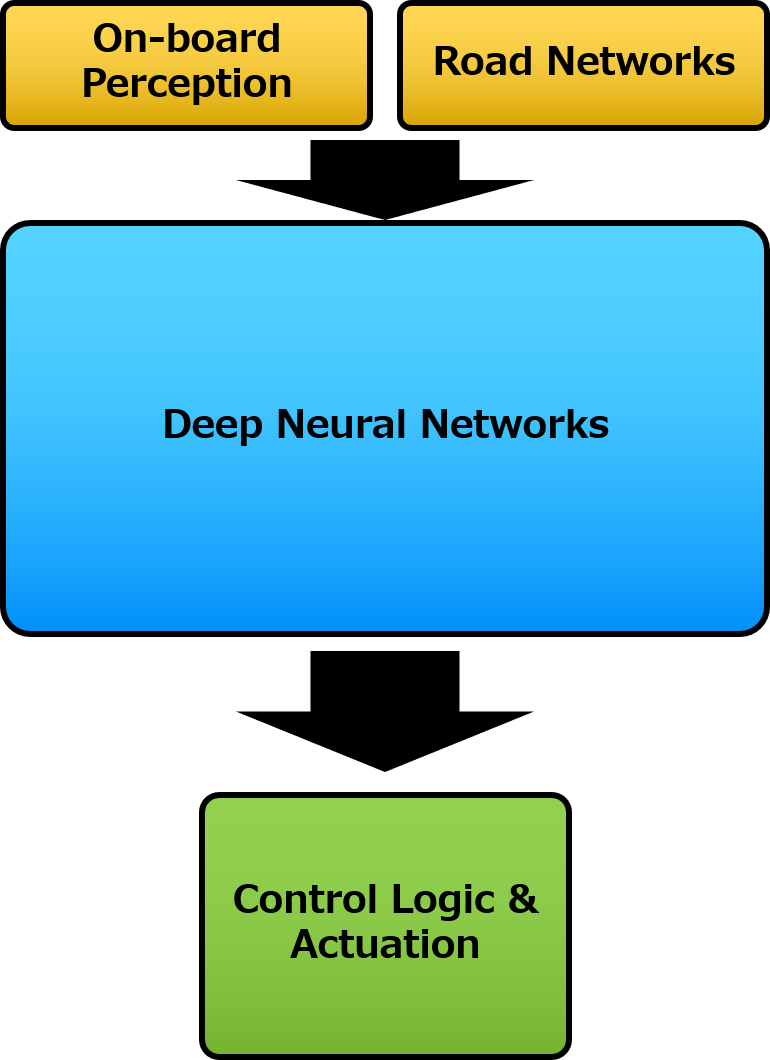}
	\hspace{2.6cm} (b) End-to-End Learning-based Approach.
          \label{fig:Synchro_3size}
        \end{center}
      \end{minipage}

    \end{tabular}
    \caption{Overall System Architecture for Autonomous Driving.}
\label{fig:E2Edriving}
  \end{center}
\end{figure}

In this paper, we present an end-to-end learning-based autonomous driving system named \textit{SuperDriver AI}, in which Deep Neural Networks (DNNs) determine the driving behaviors and policies to guarantee road safety under the mixed traffic of human-driven vehicles and automated vehicles.
In the SuperDriver AI system, we take image data as input, and outputs the values for actuating a steering wheel, a throttle, and a brake system.
SuperDriver AI learns the driving policies from the collected data of human drivers.
This approach is to envision an end-to-end learning-based autonomous driving system \cite{bojarski2016end, mori2019visual,amini2020learning}.
As shown in Figure \ref{fig:E2Edriving}-(a), the traditional autonomous driving systems \cite{urmson2008autonomous, bhat2018tools} get the data from on-board perception systems and map database, and process them for lane marking \& detection, object detection, route planning, and path planning in each module. One of the major advantages is interpretability for unexpected system behavior, and we can easily investigate the internal systems.
Above all, such autonomous driving systems are still far from complete autonomy on public roads.
As shown in Figure \ref{fig:E2Edriving}-(b), the end-to-end learning-based autonomous driving systems \cite{bojarski2016end, mori2019visual,amini2020learning} get the data for the perception systems and road networks and process them with DNNs to determine the driving behaviors.
Although such systems might be difficult to provide the interpretability, we provide a visual attention module to allow us to understand what the systems focus on and how the decision is determined.
In addition, to show the feasibility of SuperDriver AI, we collect and process driving data and test the system with real-world hardware and a vehicle.
First, we build a cloud-based data collection and learning system with the vehicle for human drivers.
Secondly, the trained networks are processed on the real-world vehicle with real-world driving scenarios.
To design and develop a robust model in real-world driving scenarios, we also develop a \textit{slit model}.

The primary contributions of this paper are as follows.

\begin{enumerate}
 \item We present a SuperDriver AI to envision an end-to-end learning-based autonomous driving system.
 \item We present a data collection and learning system for SuperDriver AI with real-world hardware and a vehicle.
 \item We demonstrate SuperDriver AI using real-world driving data and real-world hardware.
\end{enumerate}


\section{Related Works}


In autonomous driving software, as shown in Figure \ref{fig:E2Edriving}-(a), the vehicle system gets the information from the map database and the on-board perception systems that include vision cameras, LiDARs, and radars.
Also, the highly-accurate map module might be essential to such traditional autonomous driving systems.
These systems have already been deployed for factory automation and delivery robots, but they are still not deployed on public roads, mainly due to the unexpected scenarios led by human drivers and/or pedestrians \cite{tampuu2020survey, Xu_2017_CVPR}.
These traditional approaches might be called \textit{modular approach} \cite{tampuu2020survey}.

In contrast, some researchers and engineers have started to develop an end-to-end deep learning-based autonomous driving system \cite{bojarski2016end, chenJianyu2019deep, Xu_2017_CVPR}.
In such approach, the entire pipeline for autonomous driving is processed by a single neural network.
For example, NVIDIA \cite{bojarski2016end} has presented a deep neural networks-based autonomous driving with their products and they have driven a vehicle both on highways and on a smaller test field.
In addition, in \cite{chenJianyu2019deep}, the authors proposed an end-to-end autonomous driving framework with imitation learning to mimic human drivers by using simulators. They have focused on a bird-view representation as the input, and calculated vehicle trajectory as the output of the DNNs. Also, their system includes a safety and tracking controller to guarantee safety for a testing phase.
While many researchers have worked on the end-to-end approaches for autonomous driving, there are very few works using real vehicles and real-world driving scenarios, mainly because of the expensive costs.
In this paper, we design a practical imitation learning-based approach and implement it in a real-world vehicle.


\section{End-to-End Learning-based\\Autonomous Driving}

For end-to-end learning-based autonomous driving, there are two major techniques to learn driving policies: (i) imitation learning and (ii) reinforcement learning.
First, in imitation learning, a model is trained to mimic expert human drivers to learn the driving commands and timings to actuate a vehicle steering, a throttle, and a brake system.
Since the imitation learning-based autonomous driving system determines its behaviors and policies from the human driving data, the system can be difficult to navigate under traffic conditions that rarely occur.
Secondly, in reinforcement learning, a model is trained online, and the driving behaviors are determined by states and rewards.
Generally speaking, reinforcement learning is less efficient than imitation learning for learning, and it uses simulations to collect learning data because it improves driving policies from scratch.

Such learning-based autonomous driving systems might be difficult in terms of interpretability for malfunctions, but there are multiple techniques to enable us to investigate what is happening inside the models.
In fact, Kim et al. \cite{kim2019grounding} have showed a visual attention module that highlights important regions for the target image. The model predicts a spatial mask of weights and this mechanism can provide the regions where the model attends.
Autonomous driving systems can provide vehicle speed, acceleration, drivable area, and positions of the surrounding objects, by using the same inputs for the main task that determines the driving maneuvers and policies.


\section{SuperDriver AI}

\begin{figure}[!b]
\centering
\includegraphics[width=0.7\linewidth]{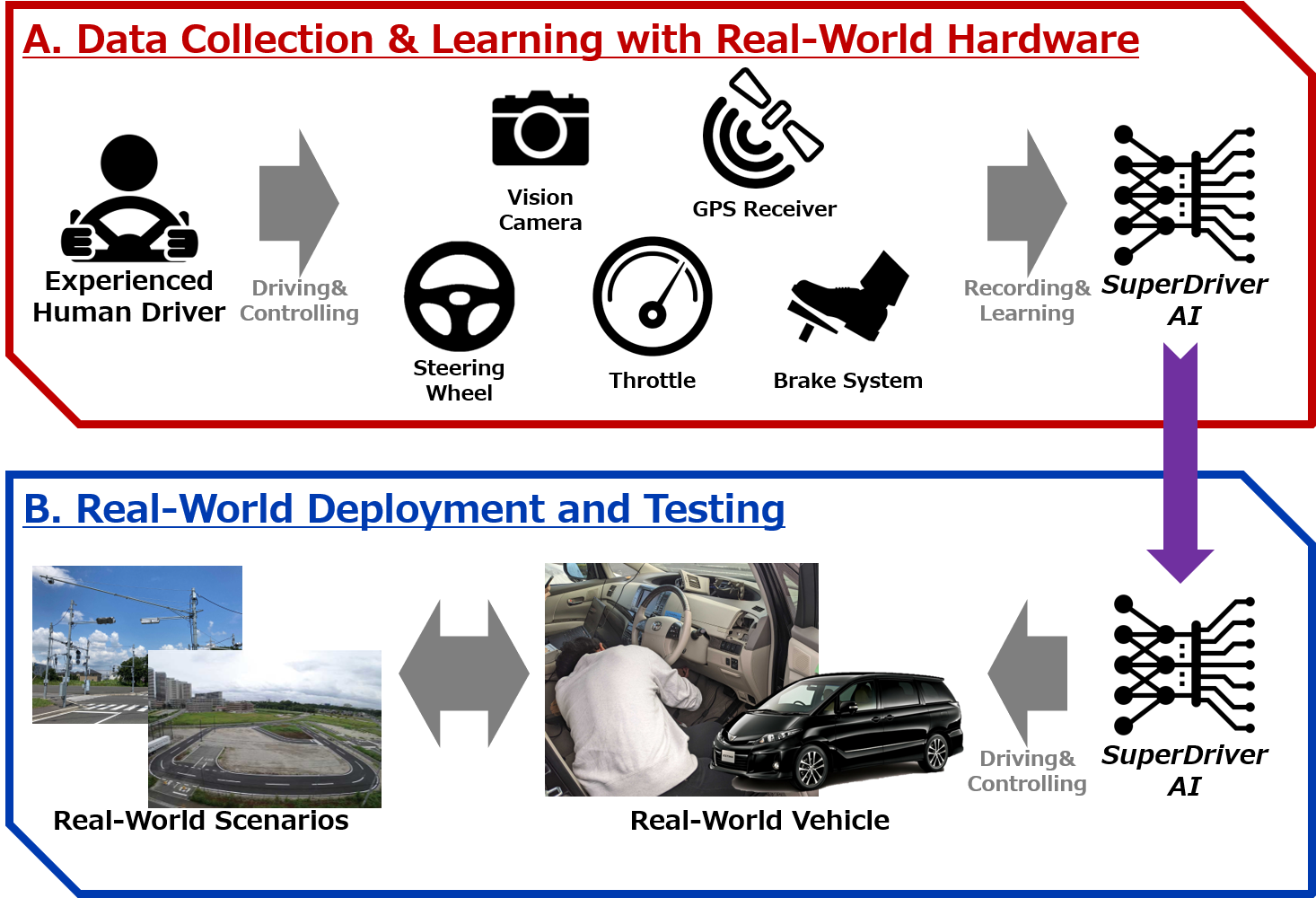}
\caption{Data Flow for End-to-End Learning-based Autonomous Driving.}
\label{fig:E2E_System}
\end{figure}

In this section, we present the overall architecture and design concepts for \textit{SuperDriver AI}, and also present \textit{slit model} to improve the robustness of the system for physical-world deployment.
SuperDriver AI is based on the imitation learning, in which we train the model with expert driving data. 
The optimal model is to produce the same driving actions and maneuvers as an expert human, by learning from the data collected from human drivers, including the data for a vision camera and a vehicle.
Hence, SuperDriver AI provides the commands for a steering wheel, a throttle, and a brake system to actuate the vehicle.
Also, we configure SuperDriver AI to mimic human drivers in terms of two factors: (i) a monocular camera and vision-only models and (ii) time-continuous inputs.
We show the feasibility of the vision-based autonomous driving systems in this paper.
Also, we understand driving conditions with the driving contexts and past moments. Likewise, SuperDriver AI needs to consider the combination of past frames for driving data.

\subsection{Architecture for SuperDriver AI}

As shown in Figure \ref{fig:E2E_System}, we have $2$ phases to train and test SuperDriver AI: (A) \textit{Data collection \& learning with real-world hardware} and (B) \textit{Real-world deployment and testing}.
To simplify the real-world implementation in these phases, we use the same hardware sets, including a vehicle and a camera.
First, to train the SuperDriver AI model, the driving data of experienced human drivers are required. In this phase, the model uses the data from an on-board vision camera, a GPS receiver, a steering wheel, a throttle, and a brake system.
Then, the trained SuperDriver AI model determines the driving actions and policies for real-world vehicles.
To drive a real vehicle with SuperDriver AI, we need to collect sufficient training data.
Also, stereo-vision cameras and/or surroundview might be useful for practical driving scenarios, but we simply use monocular camera in this paper to show the feasibility of the system.
Secondly, to test the SuperDriver AI model, a real vehicle and real-world driving scenarios might be important. 
In fact, compared to software simulation, there are many uncontrollable factors to reduce the driving quality and accuracy in the physical world. For example, the embedded cameras may be slightly misaligned during the testing phase. Also, tire friction can be changed as weather changes, and subsequently, the same input may lead to different outputs.
To enhance the robustness against such real-world factors, we present a \textit{slit model} next.

\begin{figure}[!t]
\centering
\includegraphics[width=0.7\linewidth]{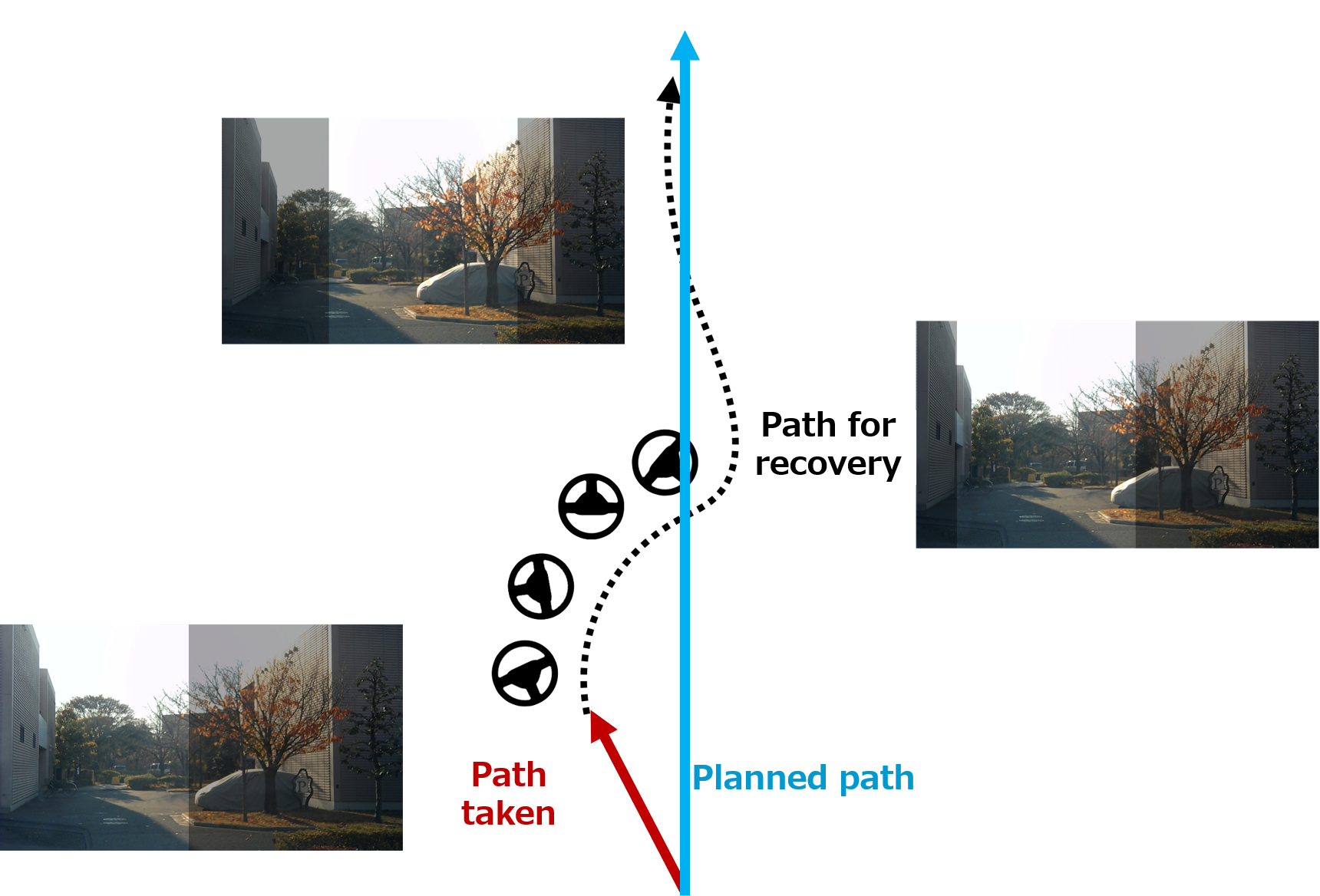}
\caption{Slit Model for Robust End-to-End Autonomous Driving System.}
\label{fig:slitmodel}
\end{figure}

\begin{figure}[!b]
\centering
\includegraphics[width=0.56\linewidth]{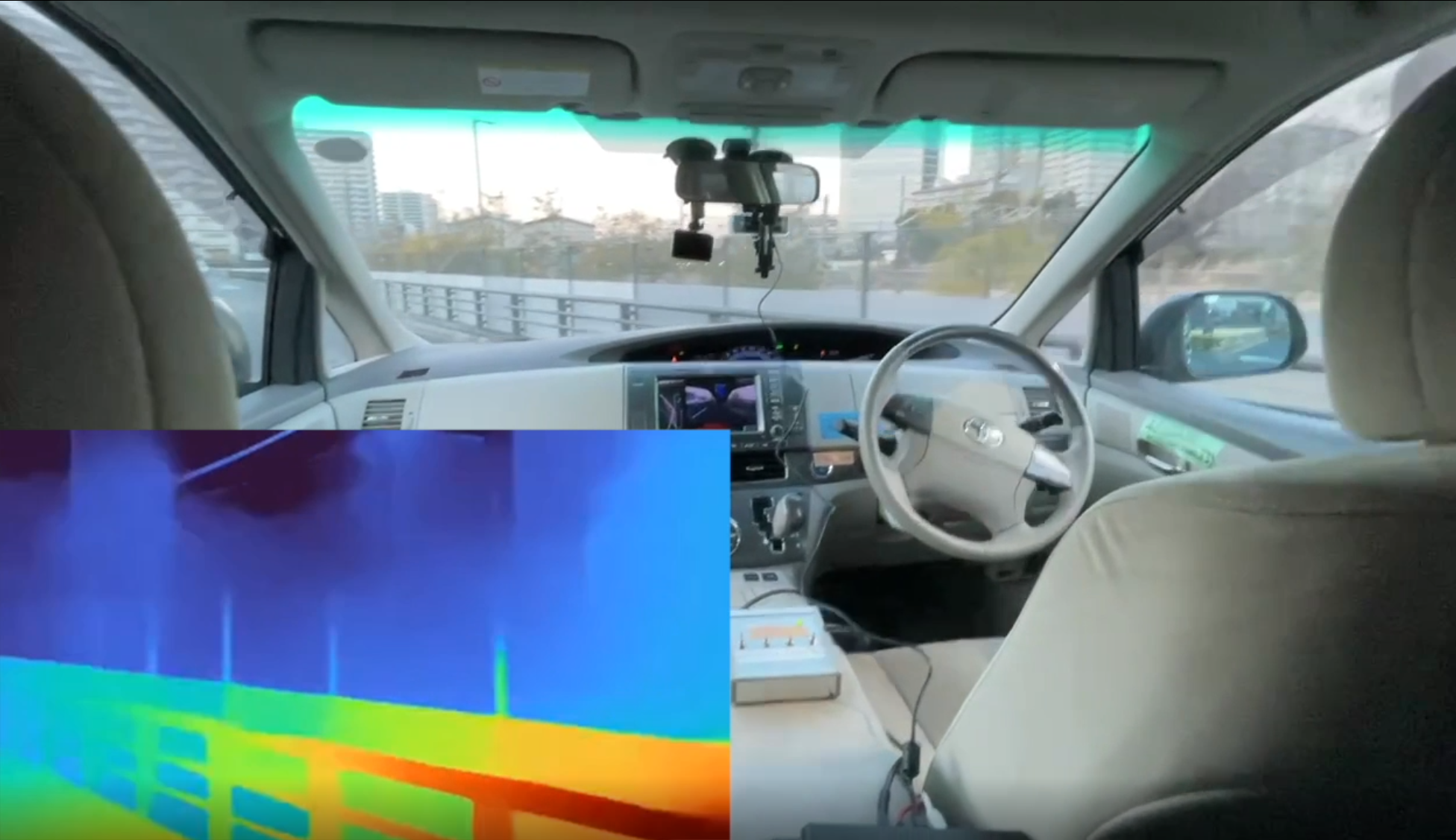}
\caption{Depth Estimation with Single Camera.}
\label{fig:singlecamera}
\end{figure}

\subsection{Slit Model: For Robust Design}

This section presents the slit model that improve the robustness of the end-to-end learning-based autonomous driving system.
The slit model is designed to tolerate the misalignment of sensor installation and the system failures/delays of real-world hardware.
In the slit model, as shown in Figure \ref{fig:slitmodel}, we crop the camera view and generate a pseudo-displacement and/or pseudo-misalignment.
Then, we train the model with such cropped views and pseudo-displacement.
By moving the cropped regions for the image data, we can train the SuperDriver AI model to recover the path taken.

\begin{figure}[!t]
  \begin{center}
    \begin{tabular}{c}
      \begin{minipage}{0.5\hsize}
        \begin{center}
          \includegraphics[clip, width=0.9\linewidth]{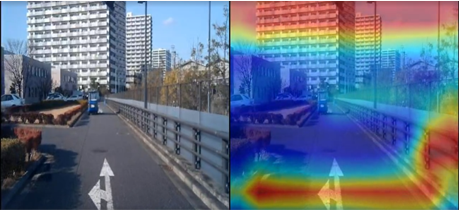}
	\hspace{5.6cm} (a) Model sees road markers.
        \label{fig:Synchro_2size}
        \end{center}
      \end{minipage}

      \begin{minipage}{0.5\hsize}
        \begin{center}
          \includegraphics[clip, width=0.9\linewidth]{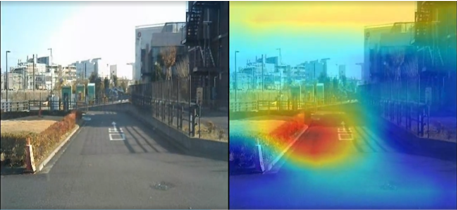}
	\hspace{5.6cm} (b) Model sees garden trees.
          \label{fig:Synchro_3size}
        \end{center}
      \end{minipage}

    \end{tabular}
    \caption{Visual Explanations for SuperDriver AI Model.}
\label{fig:gradcam}
  \end{center}
\end{figure}

As shown in Figure \ref{fig:slitmodel}, we present the overview of the slit model, including the cropped views and paths.
Here, the vehicle has slightly shifted from the original planned path and the vehicle body is off to the left in the example.
Hence, the vehicle has to recover its path by turning the steering wheel to the right.
In Figure \ref{fig:slitmodel}, the shaded regions are removed to generate the pseudo-displacement data.
When we remove the right-half of the original view, the vehicle is in a pseudo-left shifting position. We can use such data to train the SuperDriver AI model.

\subsection{Visual Attention}

Since end-to-end autonomous driving systems might be difficult to explain how the systems determine the maneuvers and what the systems focus on, the visual attention module for human riders/users are highly required.
To improve the interpretability, we implement the depth estimation module with the monocular camera embedded on the vehicle, as shown in Figure \ref{fig:singlecamera}.
As shown in Figure \ref{fig:singlecamera}, we can easily estimate the physical distance from the camera to the road blocks by using one camera and deep neural networks.
In addition, we implement Grad-CAM (Gradient-weighted Calss Activation Mapping) \cite{selvaraju2017grad} to highlight important regions in the image, as shown in Figure \ref{fig:gradcam}. In Figure \ref{fig:gradcam}-(a), the SuperDriver AI model sees the road surface to determine the driving actions. Also, in Figure \ref{fig:gradcam}-(b), the model checks the neighboring garden trees to determine actions.
In fact, by using the output of Grad-CAM, we can know that the SuperDriver AI model uses the variety of visual information to determine the driving actions.
Such visual attention modules are indirectly relevant for the autonomous driving, but these tasks provide the internal representations of the model. The human riders/users might be comfortable to know what the SuperDriver AI model focuses on in real time. Also, these tasks might be helpful to analyze the root causes when the autonomous driving systems have unexpected behavior.


\section{Real-World Deployment and Demonstration}

In this section, we present the real-world data collection, deployment, and implementation.
We have implemented the SuperDriverAI system, and for safety purposes, we keep the vehicle speed constant by actuating only the steering wheel.

\begin{figure*}[!t]
\centering
\includegraphics[width=0.6\linewidth]{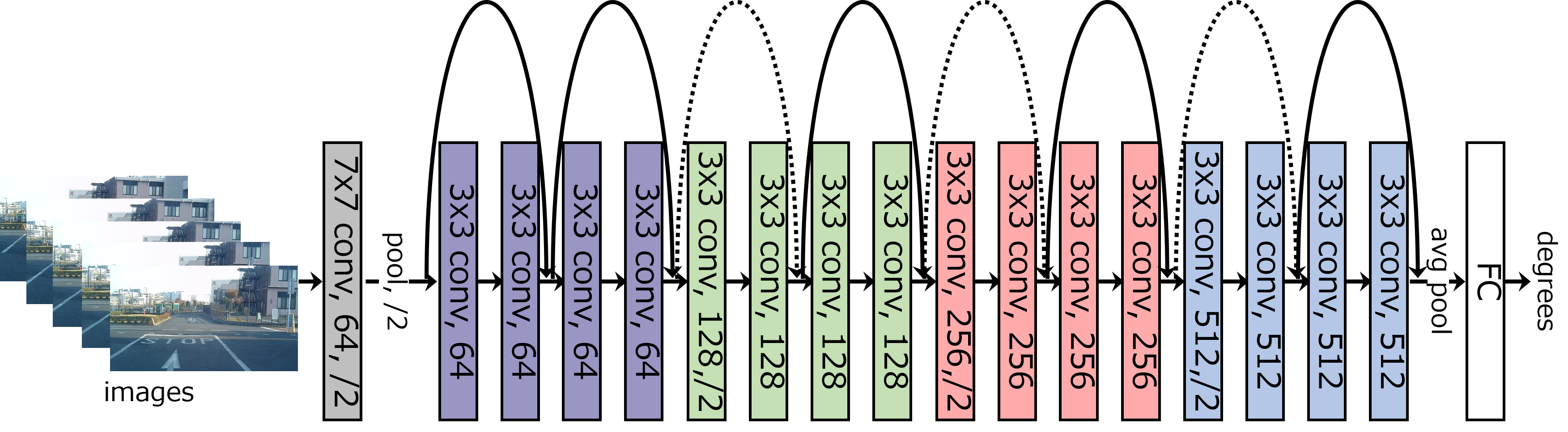}
\caption{DNN Architecture for Autonomous Driving.}
\label{fig:dnn_architecture}
\end{figure*}

\subsection{Data Collection and Deployment}

As shown in Figure \ref{fig:E2E_System}, we have collected the variety of data from experienced human drivers, and we have used such data for autonomous driving in practice.
We have developed a cloud-based data platform to automatically upload the driving data and to download the SuperDriver AI model. Hence, once the human driver completes the driving, the data are uploaded to the cloud server without effort.

In addition, we use \textit{Toyota Previa} to collect the driving data and to deploy our SuperDriver AI model.
We have implemented the hardware tools to collect the data from the vehicle system, and we also have an embedded vision camera to collect the time-continuous image data.
Also, to test the vehicle, we have used a practical test field at Kawasaki City, Japan.
We have collected 150 runs for one driving scenario that involves both straight and turning.

\begin{figure}[!b]
\centering
\includegraphics[width=0.76\linewidth]{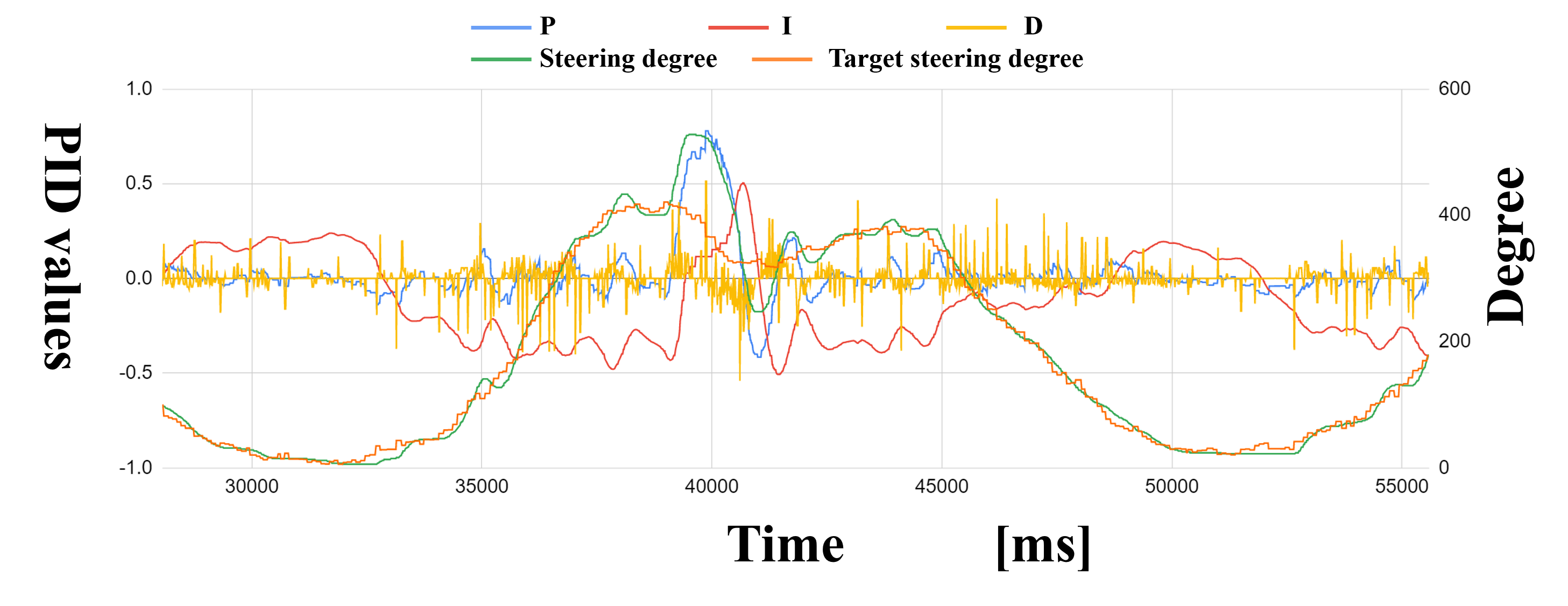}
\caption{Data from Vehicle Systems for Actuation.}
\label{fig:CAV_realdata}
\end{figure}

\subsection{Implementation \& Evaluation}

In this section, we present the DNN implementation and real-world demonstration with the real vehicle.
First, we show the DNN architecture for the SuperDriver AI model, as shown in Figure \ref{fig:dnn_architecture}.
The baseline model is ResNet18 \cite{he2016deep} that is one of the most prevalent architecture for deep neural networks.
Since the driving tasks are timely-continuous ones, the network models use multiple frames ($t=1...n$) as input data, which is approximately for 3 seconds. In addition, the model outputs the values for steering wheel as a angle for several time steps.

For the evaluation of the system, we show the data from the vehicle systems when we use the SuperDriver AI model, as shown in Figure \ref{fig:CAV_realdata}.
Figure \ref{fig:CAV_realdata} presents the values for PID (Proportional-Integral-Derivative) controller, for an actual steering degree, and for a target steering degree.
As shown in the figure, the desired value and actual value are sufficiently close.
This means that the vehicle actuation is accurately conducted.
The vehicle contains the multiple computers to run the SuperDriver AI model in real time.
Here, one concern might be hardware flexibility and diversity.
In fact, through the learning and testing phases, we use the same hardware sets to show the feasibility of SuperDriver AI, but the sensory values might be different based on the hardware.


\section{Conclusion}

In this paper, we presented an end-to-end learning-based autonomous driving system named \textit{SuperDriver AI}, where Deep Neural Networks (DNNs) determine the driving maneuvers while ensuring road safety.
To improve robustness and interpretability, we also designed and developed a visual attention module.

In future work, we will extend the imitation learning-based autonomous driving systems for more practical scenarios, in order to show the scalability of the end-to-end-based approaches.
In addition, we will study and test the SuperDriver AI model to understand the driving policies and intentions of the surrounding vehicles, in order to enable safe cooperation between autonomous vehicles and human-driven vehicles.
Since we might require a long transition period to replace all vehicles with autonomous vehicles, such abilities would become important in the mixed traffic environments.
















\bibliographystyle{./bibliography/IEEEtran}
\bibliography{reference}

\end{document}